%% file: main.tex
\documentclass[sigconf,nonacm]{acmart}

\setcopyright{none}
\renewcommand\footnotetextcopyrightpermission[1]{}
\usepackage{graphicx}
\usepackage{booktabs}
\usepackage{array}
\usepackage{xspace}
\usepackage{xcolor}
\usepackage{tikz}
\usetikzlibrary{arrows.meta,positioning}

\newif\ifshowappendix
\showappendixfalse

\newcommand{\appnote}[1]{\ifshowappendix{} (#1)\fi}

\newcommand{\sys}{\textsc{AgentPProf}\xspace}
\newcommand{\operation}{\texttt{operation}\xspace}
\newcommand{\opstack}{\texttt{operation stack}\xspace}

\begin{document}

\title{AgentPProf: Semantic Profiler for Long Horizon AI Agents}

\author{%
Yusheng Zheng$^{1,2}$ \quad Chaokun Chang$^{3}$ \quad Yu Mao$^{4}$ \quad Tianyuan Wu$^{3}$ \quad Yuxi Huang$^{2}$ \\
Tao Ma$^{5}$ \quad Wenan Mao$^{5}$ \quad Shuyi Cheng$^{5}$ \quad Andi Quinn$^{1}$ \quad Wei Wang$^{3}$ \\[4pt]
{\normalsize $^{1}$UC Santa Cruz \quad $^{2}$Eunomia Labs \quad $^{3}$HKUST \quad $^{4}$Independent Researcher \quad $^{5}$Alibaba Group}
}

\renewcommand{\shortauthors}{Zheng et al.}

\begin{abstract}
AI agents increasingly orchestrate long-running activities with users, tools, and system resources for days and weeks.
To improve agent quality, safety, and cost efficiency, developers need to determine where failures happen, what triggers unsafe effects, and which tasks consume the most budget, then optimize those tasks.
In systems software, profiling answers similar questions by aggregating resource consumption and attributing it to responsible code paths to identify hotspots.
Yet existing agent observability tools focus on per-execution debugging and tracing rather than cross-run, long term profiling, making these questions difficult to answer at scale.
Agent observability needs profiling, not only debugging, but profiling agents is challenging: the responsible entities are task intent like \emph{diagnose authentication}, \emph{compare branches} rather than code paths, and lack stable identifiers for aggregation.
We propose a \emph{semantic operation stack model} that adapts profiling to agent trajectories.
Uniform \emph{operations} represent all activities, and \emph{operation stacks} replace the runtime call stack, enabling hierarchical attribution at different granularities.
We observe that an agent's task occupies a contiguous span and decomposes into subtasks, so we introduce \emph{recursive operation segmentation}, which recursively splits trajectories at task boundaries.
\sys is a profiler that aggregates agent trajectories into pprof-compatible profiles, enabling flame graph visualization and analysis.
\sys reaches 0.764 B$^3$ F1 against human annotations on CodeTraceBench. On three problem-localization benchmarks, the profile raises MAP by up to 56\%, demonstrating that it effectively attributes resources, locates problems, and helps optimize token cost at practical profiling cost. \sys is available at \url{https://github.com/eunomia-bpf/agentsight}.
\end{abstract}

\maketitle




\section{Introduction}
\label{sec:intro}

\begin{figure}[!t]
\centering
\includegraphics[width=0.84\columnwidth]{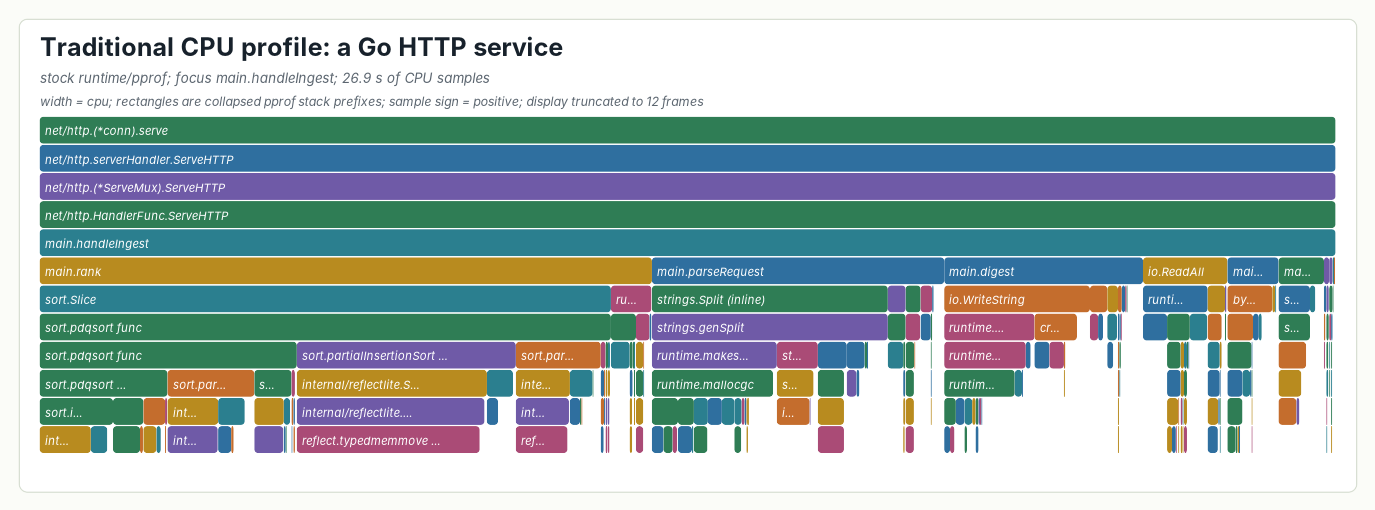}\\[1pt]
\includegraphics[width=0.84\columnwidth]{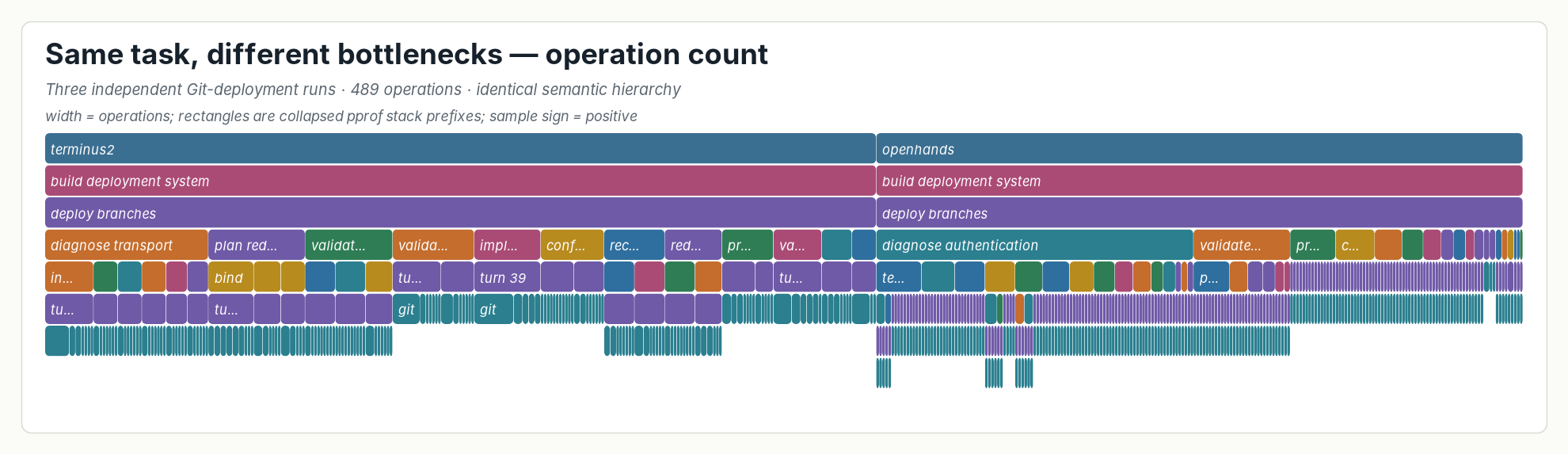}\\[1pt]
\includegraphics[width=0.84\columnwidth]{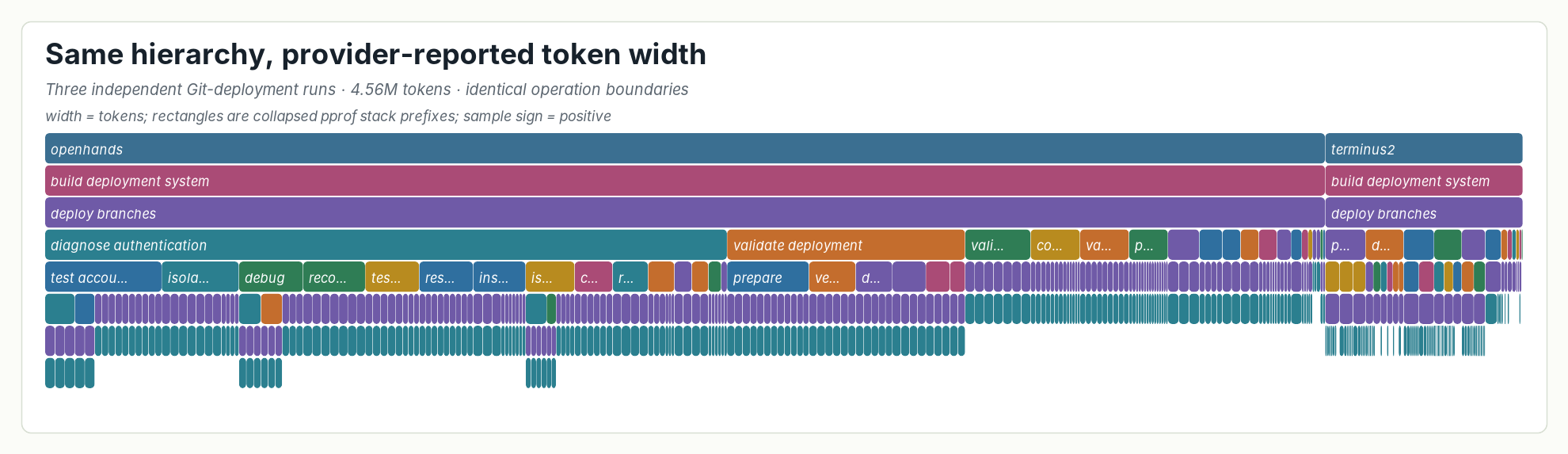}\\[1pt]
\includegraphics[width=0.84\columnwidth]{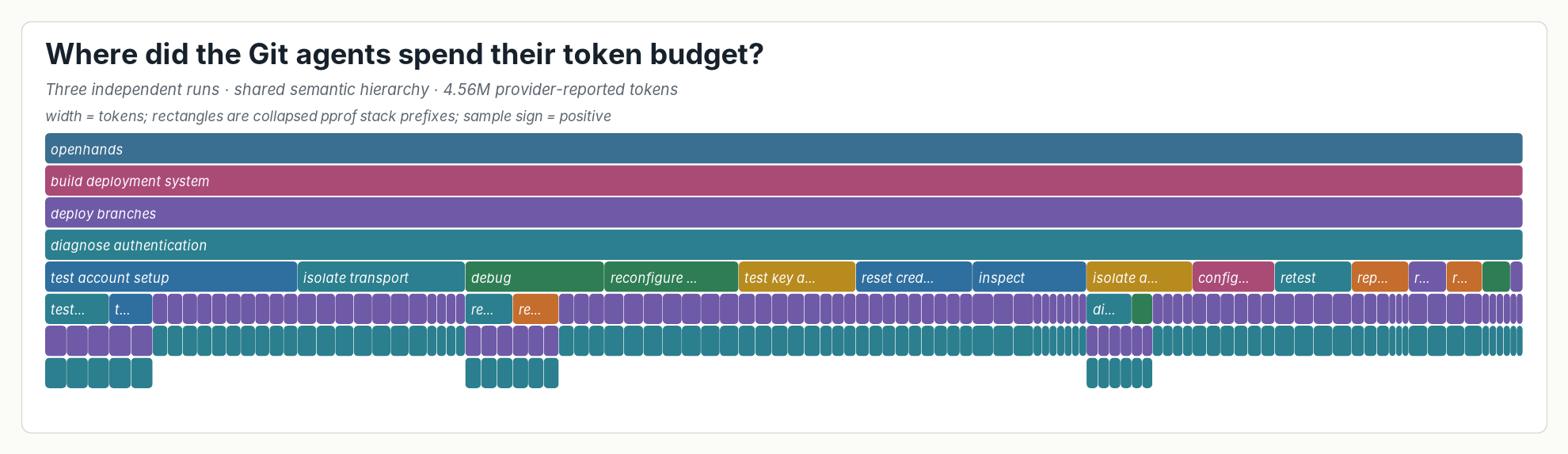}
\caption{One renderer, four standard pprof profiles.
\emph{(a)} CPU time in a Go HTTP service.
\emph{(b, c)} Three agent executions under one fixed hierarchy: width is operation count (b) or tokens (c).
\emph{(d)} Token profile focused on \texttt{diagnose authentication}.}
\Description{Four flame graphs in one visual style. The first descends from a
Go HTTP server's connection handler into request handling, ranking, parsing,
and hashing. The second and third show the same agent hierarchy weighted by
operation count and by tokens, where the count view is dominated by one agent
framework and the token view by another together with repeated
authentication diagnosis. The fourth expands that authentication
responsibility into account, transport, credential, and retest operations.}
\label{fig:flamegraph}
\end{figure}

AI agents~\cite{sweagent,claudecode,osworld} orchestrate multi-step activities spanning two layers: high-level intent (prompts, LLM calls, tool invocations) and low-level system effects (process spawns, file and network I/O).
As agents evolve from short scripted workflows into complex systems that run for hours or days, production teams accumulate many such trajectories over weeks or months.

To improve agent quality, safety, and cost efficiency, developers
need to analyze operational behavior across trajectories and
interactions.
Which categories of tasks consume the most token budget? Where do
failures concentrate across workflows? Which behavioral patterns
trigger unsafe system effects?
Answering these questions is the first step toward optimizing high-cost tasks and fixing failure-prone workflows.
Answering them requires analysis and evaluation across many
trajectories~\cite{agentatlas,agentrewardbench}, which is increasingly costly for manual inspection or
per-trajectory LLM evaluation~\cite{llm-as-judge} at scale.
In traditional systems software, profiling answers these questions by
aggregating resource consumption and attributing it to responsible
entities, distinct from per-execution debugging and tracing.

Yet existing agent tools support debugging and tracing but not
profiling.
Some~\cite{langsmith,langfuse,phoenix,otel} organize events into
per-execution span trees, effective for diagnosing a single run but
unable to aggregate by task intent or workflow phase.
Others~\cite{datadog-llmobs,laminar-signals} cluster inputs or
extract structured signals, but characterize \emph{input
distributions} (``30\% of users ask code questions'') rather than
\emph{resource attribution} (``review tasks consume 40\% of the
token budget''), because request-level tags do not propagate
to downstream tool calls and system effects.

Adapting profiling to agent trajectories is challenging: in
traditional software, the responsible entities are code paths, and
profiling is straightforward because function names are stable and
runtime call nesting provides the attribution hierarchy.
Agent behavior differs. To answer the questions above, the responsible entities must be task intent and workflow phase, not code paths.
Agent trajectories lack the two properties that make traditional profiling possible.
Prompts are natural language, so two prompts expressing the same intent share no common string.
Agent events have no runtime call-stack hierarchy because prompts,
tool calls, and system effects are not nested by execution the way
function calls produce stack frames. A sequence of prompts can be a task or multiple tasks, and a task can be part of a bigger task.

Despite these differences, the fundamental profiling methodology
transfers to agent trajectories by projecting resources onto responsible
entities, assigning stable identifiers, and attributing hierarchically.
We propose a \emph{semantic operation stack model} with two
components.
\emph{Operations} are uniform records with string fields and
additive measures representing all agent activities (prompts, LLM
calls, tool invocations, system effects).
\emph{Operation stacks} provide hierarchical attribution, replacing the runtime call stack.
Choosing different fields changes the attribution granularity: the same operations can be attributed by task category, by execution phase, by session, or by action type, and one fixed hierarchy replays under any additive measure (Figure~\ref{fig:flamegraph}).

We implement this model in \sys, an offline profiler that compiles
local agent trajectories into pprof-compatible profiles
(Figure~\ref{fig:architecture}).
One algorithm, \emph{recursive operation segmentation}, supplies the missing identifiers.
Labeling every prompt individually would be prohibitively expensive, but task structure is recoverable at lower cost: a task occupies a contiguous span of a trajectory and decomposes into contiguous subtasks, so recovering it requires only finding the transitions between them.
The algorithm finds where responsibility changes, names the resulting intervals, and yields short names starting with a verb (like \texttt{diagnose authentication}) that fold across sessions. The interface is implementation-neutral.

We evaluate whether \sys effectively attributes resources, locates problems, and recovers human-recognizable task structure at practical cost.
Across all 405 CodeTraceBench~\cite{li2026codetracer} trajectories, recursive segmentation agrees with human stage annotations at 0.764 B$^3$ F1~\cite{bagga-baldwin-1998-entity-based}, versus 0.663 for a statistical baseline and 0.541 for raw actions.
On three complete localization benchmarks, the profile improves ranking over each benchmark's own diagnostic, raising MAP by 0.031, 0.107, and 0.117; as a navigation aid, it reduces content to inspect by 12 percentage points at equal ranking quality.
A profile-derived repair reduces agent tokens by 19\% without degrading task quality, and annotation cost is reduced by 20\%.

This paper makes three contributions:

\begin{enumerate}
\item \textbf{Semantic operation stack model.}
  We identify the missing profiling layer in agent observability and
  propose a model of uniform operations and query-time operation
  stacks (Background and Design).
\item \textbf{System: \sys.}
  A profiler that implements this model, producing pprof-compatible output.
\item \textbf{Evaluation.}
  We evaluate profiler output against independent ground-truth
  annotations that stay hidden from the profiler, on eight public
  benchmarks and three real trajectory datasets.
\end{enumerate}

\section{Background and Motivation}
\label{sec:background}

\paragraph{LLMs and AI Agents.}
A typical agent trajectory interleaves two layers of activity.
At the intent layer, the agent receives a user prompt, issues LLM
calls, and invokes tools (code execution, web search, file editing).
Each tool invocation triggers system effects at the lower layer:
process spawns, file reads and writes, and network requests.
A single trajectory may contain hundreds of cycles between intent
and system effects.

\paragraph{System Profiling.}
In traditional software, a profiler periodically samples execution state, attaches each
sample to the current call stack, and merges samples with identical
stacks.
Flame graphs~\cite{flamegraphs} and pprof~\cite{pprof} call graphs
visualize the result, where width or node size represents aggregate
cost.
Engineers use these profiles to find where CPU or memory budget is spent, then optimize the responsible code paths to reduce cost.
These tools support flexible aggregation, but they all assume
stable function names and a runtime call stack.
Pprof's \texttt{tagroot}/\texttt{tagleaf} options can add
label-derived pseudo-frames, but only on top of an existing execution
stack that agent trajectories do not have.

\paragraph{A Motivating Case.}
Three independent agent attempts at one real Git-deployment task all failed to deliver the requested password-authenticated endpoint.
The developer's question is \emph{what went wrong, and is it the same problem across all three runs?}
Reading three transcripts totaling 489 operations and 4.6M tokens would take hours.
Figure~\ref{fig:flamegraph} places a conventional CPU profile beside the
semantic profile of those three runs.
Both are standard pprof profiles drawn by one renderer and read by identical
rules, where width is aggregate cost, a parent contains its children, and identical
stacks recorded at different moments fold into one frame; what differs is where the frames come from: a call stack supplies \texttt{net/http.(*conn).serve} at every sample for
free, but nothing in an agent trajectory supplies \texttt{diagnose
authentication}, because the three runs express that intent through different
prompts and shell commands.

The profile answers in one view (Figure~\ref{fig:flamegraph}): all three runs share a \texttt{diagnose authentication} responsibility that consumes 46\% of their combined token budget but only 21\% of operation count.
Expanding the view shows all three executions verifying substitute transport paths without ever establishing the target endpoint: the agents kept retrying SSH variants instead of fixing the actual credential issue, and operation count alone hides this because authentication commands are cheap to issue but expensive to verify.
The profile points to where to intervene: early credential validation or bounded retry depth in \texttt{diagnose authentication} could cut token cost for similar tasks.

\paragraph{Challenges for Agent Profiling.}
Adapting profiling to agent trajectories faces two core challenges.
First, the responsible entities differ: cost must attach to task intent and workflow phase rather than code paths, and existing tools~\cite{otel,openinference} do not capture these entities because prompts are natural language---two prompts expressing the same intent may differ, so the profiler cannot aggregate them the way it aggregates identical function names.
Second, agent trajectories have no runtime hierarchy for attribution: in CPU profiling the call stack fixes attribution at every level (\texttt{malloc}'s cost belongs to \texttt{parse}, to \texttt{process}, and to \texttt{main} simultaneously), but a sequence of prompts may constitute one task or several, a task may be part of a larger workflow, and no runtime mechanism marks these boundaries or determines the attribution granularity.

\section{Design}
\label{sec:design}

Agent profiling needs the three mechanisms that call stacks provide automatically, but must achieve them without a call stack:
\textbf{R1} (cross-layer projection): connect system-layer consumption to intent-level categories;
\textbf{R2} (stable identifiers): derive short, repeatable identifiers from natural-language prompts before folding;
\textbf{R3} (hierarchical attribution): construct attribution hierarchies from data, not execution nesting.

A view is a triple $(\varphi, \sigma, w)$: a predicate
$\varphi$ selects which operations participate, a stack function
$\sigma = [f_1, \ldots, f_k]$ maps each operation to its frame
sequence, and a weight function $w$ assigns a nonnegative measure.
Four built-in views cover common questions: \texttt{tokens} (LLM calls weighted by token count), \texttt{time} (all timed operations weighted by duration), \texttt{files} (file operations weighted by event count), and \texttt{network} (network operations weighted by event count).
Operations satisfy R1 by representing intent and system-effect activities in a single record with tag inheritance, so intent tags propagate to the system effects they trigger via AgentSight~\cite{agentsight}.
\emph{Recursive operation segmentation} satisfies R2 and R3 together, deriving short, stable interval names from trajectory content and nesting them into the attribution hierarchy that replaces the runtime call stack.
\emph{Operation stacks} then attribute resources at every level of that hierarchy at query time, so the same data supports multiple views without rebuilding the input.
The profiling pipeline has four stages: (1)~parse raw trajectories into operations, (2)~segment trajectories into named nested intervals (or apply mapping rules), (3)~project each operation onto a stack frame sequence chosen at query time, and (4)~fold operations with identical stacks, summing their weights.

\subsection{Semantic Operation Stack Model}
\label{sec:ops}

This section defines operations and operation stacks, the data structures that satisfy R1 (cross-layer projection) and enable R3 (hierarchical attribution). R2 (stable identifiers) requires deriving identifiers from trajectory content and is addressed in Section~\ref{sec:intent}.

Because agent activities span both intent and system-effect
layers, a profiler should not distinguish them in the
data representation: \sys represents every activity as a single abstraction, the
\operation, a weighted record with string fields and additive
measures.
Each operation carries string fields (\texttt{project},
\texttt{agent}, \texttt{session}, \texttt{prompt\_tag},
\texttt{kind}, \texttt{model}, \texttt{path}, \texttt{domain},
\texttt{status}, \ldots) and additive measures (token count,
duration, event count).
A prompt, an LLM call, a tool invocation, a file read, a GUI click,
and a process event are all operations with the same schema, distinguished only by which fields carry values.
Only source content, timestamps, and additive measures are required; explicit tool/event IDs, sessions, and spans are optional.
\sys uses recorded identifiers verbatim when present, falls back to the lifetime-overlap rule above when they are missing, and leaves ambiguous or conflicting links unassigned (fail closed) rather than inferring a parent; only the semantic interval names are inferred, derived from content by segmentation.

\label{sec:stacks}
An \opstack provides hierarchical attribution for agent trajectories, replacing the runtime call stack. An ordered list of fields
$[f_1, f_2, \ldots, f_k]$ projects each operation $o$ to a frame
sequence $\langle o.f_1, o.f_2, \ldots, o.f_k \rangle$, and
operations with identical sequences are merged, summing their
weights.
Unlike a call stack, the fields are chosen at query time, not
determined by execution.
Changing the fields changes the attribution without changing the data, so the same operations can be attributed by task, by session, or by action type.
Sessions and spans are optional fields, so the same data supports both debugging (include \texttt{session}) and aggregate profiling (exclude it).

\begin{figure}[t]
\centering
\resizebox{\linewidth}{!}{\input{figures/fig-architecture.tex}}
\caption{Data-flow pipeline. Local histories and public datasets both produce uniform operations; segmentation runs once, projection and folding at query time.}
\Description{Data-flow diagram from agent trajectories through parsing,
operation segmentation, and projection to folded-stack output.}
\label{fig:architecture}
\end{figure}
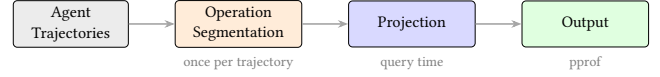

\subsection{Recursive Operation Segmentation}
\label{sec:intent}

Operation stacks can project any field an operation already carries, but agent trajectories lack the two fields that make traditional profiling work: stable task identifiers and a nesting hierarchy. Recursive operation segmentation derives both fields from trajectory content at lower cost by marking only the transition points where responsibility changes.

The algorithm rests on one structural intuition: a task occupies
a contiguous span of the trajectory and decomposes into smaller
contiguous subtasks, so we model task structure as a set of nested named intervals and recover it by finding the transition points between them.

The algorithm reads a trajectory, finds the transition points where the agent's responsibility changes, and names the intervals on both sides.
It then recurses into each interval, cutting again wherever a
finer responsibility change remains, and stops when an interval reads as
one coherent piece of work.
Formally, a trajectory is an ordered step sequence $T=(t_1,\dots,t_n)$,
and a segmentation is a set $S$ of named intervals over $T$ that is
nested (any two intervals are disjoint or one contains the other),
covers every step, and contains the session-wide interval as its root.
The algorithm computes $S$ by one recursive rule: $\textsc{Segment}(I)$ selects zero or more transition points inside interval $I$, which partition $I$ into consecutive child intervals. Each child receives a short name starting with a verb and is segmented recursively. Selecting no transition terminates the branch.
A step's operation path is the name chain of the intervals containing it, from the session root to its innermost interval. Equal paths fold across sessions.
Concretely, in one Git-deployment execution the algorithm cuts the session into \texttt{build deployment system} and, inside it, cuts again where the agent stops writing configuration and starts debugging access, yielding \texttt{deploy branches} and \texttt{diagnose authentication}. A step inside the latter carries the path \texttt{build deployment system > diagnose authentication}, and because the same names reappear in the other two executions, their costs fold together in Figure~\ref{fig:flamegraph}.
Any implementation able to find transitions and name intervals (an agent, a language model, or a statistical rule) can serve as the segmenter.
In evaluation, we use Codex~\cite{codex} with GPT-5.6-sol-high, with one fixed instruction per trajectory:

\begin{quote}\small
\emph{Input:} one trajectory (each step's prompt, command, and output summary, with no labels or scores).\\
\emph{Action:} read it, emit a mark wherever the responsibility changes, and use \sys to check and update the segmentation, and re-read and update the marks iteratively until you think the segmentation is complete.\\
\emph{Output:} sparse path marks, e.g.,\\
{\footnotesize\ttfamily step 1: [deploy git server]\\
step 4: [deploy git server, diagnose SSH access]\\
step 15: [deploy git server, verify web endpoint]}\\
---one line only where the path changes; names start with a verb, one to three words.
\end{quote}

The sparse marks still produce a complete segmentation because steps between two marks inherit the preceding path.
Segmentation is iterative, so the agent can revise marks incrementally without re-annotating the entire trajectory.
The model supports both offline evaluation and online annotation during execution.

\section{Implementation}
\label{sec:impl}

\sys is a Rust CLI ({$\sim$}9.8K LOC, Figure~\ref{fig:architecture}) that realizes the design of Section~\ref{sec:design}.
It reads Codex and Claude Code local JSONL history files and AgentSight~\cite{agentsight} recordings for system effects. These files are written incrementally during agent execution, so \sys can build profiles while the agent is still running.
The pipeline has two stages: (1) preprocessing extracts a readable trajectory summary, and (2) the segmentation agent reads this summary and writes interval marks to an annotation file, which it can revise until satisfied.
The CLI validates that marks are nested and cover every step, then emits a standard pprof protobuf profile that \texttt{go tool pprof} reads directly.
Projection is deterministic: each operation contributes its semantic path and LLM/tool evidence, with session identities kept as pprof labels rather than visible frames, so equal semantic prefixes fold across sessions.
Switching the additive measure replays the same annotations and changes only stack widths.

Fields already recorded literally (e.g., skill invocations) become stack levels without annotation cost.
Multi-agent orchestration works automatically because subagent operations inherit the outer agent's semantic path.
A non-LLM statistical segmenter is also available, scoring transitions by NPMI~\cite{bouma2009npmi} and calibrating an unsupervised cutoff with $k$-means~\cite{macqueen1967}.
\section{Evaluation}
\label{sec:eval}

We evaluate whether \sys effectively attributes resources, locates problems, and recovers human-recognizable task structure at practical cost. The evaluation uses three data classes.
\emph{Real inputs}: 41 long-horizon coding-agent sessions (three of which
repeat one Git-deployment task and form the RQ1 case), 440 mixed
web-agent trajectories~\cite{agentrewardbench}, and one developer's complete
local session history from the authors' workstation (1{,}394 sessions), whose
42 long-horizon development sessions are the annotated portion.
\emph{Annotated trajectories}: 405 CodeTraceBench trajectories with
20{,}866 operations and 2{,}948 human-annotated
stages~\cite{li2026codetracer}, 287 OSWorld-Human
sessions~\cite{osworldhuman}, 1{,}012 AgentBoard goals~\cite{agentboard},
and 2{,}737 published action
labels~\cite{bouzenia-pradel-2025-trajectories}.
\emph{Problem-localization benchmarks}: the complete AgentProcessBench,
HINTBench, and TraceElephant workloads, with independently annotated faulty
operations, over 27{,}346
operations~\cite{agentprocessbench,hintbench,traceelephant}.
All annotations, outcomes, and scores stay hidden from every method
until its output is produced.
Benchmark trajectories average 8--52 operations (short to medium); the 42-session workstation portion supplies long-horizon coverage.

\subsection{RQ1: Does Semantic Profiling Improve Resource Attribution?}

\emph{Setup.} Improved resource attribution should (1) reunite cross-run responsibility scattered by alternative organizations and (2) reveal different bottlenecks across additive measures.
We test both on 41 real long-horizon agent trajectories (3{,}146 user turns, 5{,}750 operations), including three independent runs of one Git-deployment task whose 735 steps receive 96 recursive annotations to semantic depth five.

\paragraph{The motivating case revisited.}
The Git-deployment case in Section~\ref{sec:background} shows semantic organization reuniting responsibility scattered by alternative organizations; here we add elapsed time as a third measure.
The hierarchy replays under all three measures with exact conservation, while \texttt{diagnose authentication}'s share rises from 21\% (count) through 37\% (time) to 46\% (tokens). Changing only the measure more than doubles the attributed share.
A quantitative control shows why both alternatives fail.
\emph{Raw-action grouping} (by literal action-type, e.g., \texttt{run}, \texttt{edit}) fails: 102 of 105 authentication operations are labeled only \texttt{run}, whose bucket includes 97 unrelated operations. Coarser action-kind grouping scatters them across six kinds (execute 39\%, version-control 22\%, edit 19\%, inspect 12\%, search 7\%, install 1\%).
\emph{Native call trees} place every operation under its distinct source LLM/tool call, without a shared responsibility node.
Only the semantic hierarchy reunites authentication into one focusable cross-run path while preserving all call/tool evidence as leaves.
Separately, real AgentSight~\cite{agentsight} eBPF recordings likewise fold
all 1{,}520 captured process spawns and file operations under task
responsibilities with exact conservation, demonstrating end-to-end system effects.

\paragraph{Use case 2: where did this project's agent budget go?}
A developer spent weeks building \sys{} and this paper with coding agents; we annotate the history's long-horizon portion (42 sessions: 18 Codex and 24 Claude Code; 1{,}252 prompts, 5{,}620 LLM calls, and 1.38 billion tokens).
The question is \emph{what did the agents actually spend tokens doing?}
The profile shows a broad distribution without a dominant sink; the largest path (\texttt{refine paper > align evaluation}) holds only 1.7\% of tokens.
The three longest sessions (each tens of hours) have distinct dominant responsibilities: evaluation alignment, evidence inspection, and merge resolution.
Switching to operation count shifts the concentration. Token mass stays at prompt depth (70\%), whereas operation mass resolves deeper (44\% at depths three and four).
The history also demonstrates scaled multi-agent composition: 98 subagent delegations across 14 sessions, each containing all downstream subagent operations and composing with annotated semantic levels.

The hierarchy replays under FILE-READ, FILE-WRITE, and NETWORK-target widths, making side effects auditable by responsibility.

Literal skill invocations require no annotation, thus covering the complete history (1{,}394 sessions, 6.9B tokens).
There, \texttt{paper-writing-style} leads both measures (29M tokens, 99.47\% cache reads), making its repeated-context workflow the first inspection target. The largest session holds only 31\% of that mass, so this priority appears only after folding.
The 99.47\% cache-read ratio suggests repeated rebuilding of similar context. Reusing context across invocations could reduce token cost.

Across 440 AgentRewardBench trajectories (7{,}229 operations, 51{,}904{,}621 tokens, both exactly conserved), count- versus token-based rankings agree strongly (mean tau-b 0.886).
Multi-measure profiling adds value precisely for the 13\% of tasks below 0.7 agreement\appnote{details in Appendix~\ref{app:rq1}}.

\emph{Takeaway.} Both hold: the semantic hierarchy reunites responsibility scattered by raw-action and native organizations (authentication), while switching measures on that hierarchy changes attributed share by over 2$\times$ (21\% to 46\%), exposing bottlenecks hidden by operation count. These optimization targets can be restructured to reduce disproportionate task costs.

\subsection{RQ2: Does Profiler Output Correspond to Real Problems?}

\emph{Setup.} Three tests assess correspondence between profiler output and independently annotated real problems: (1) \emph{differential analysis} compares profile failure structure with expert behavioral labels across 440 runs; (2) \emph{localization} tests whether profile scores improve existing benchmark judges' fault rankings; (3) \emph{reading study} tests whether semantic names preserve an LLM agent's ranking with less inspection.

\begin{figure}[t]
\centering
\textbf{(a) Recovery focus}\\[-2pt]
\includegraphics[width=\columnwidth]{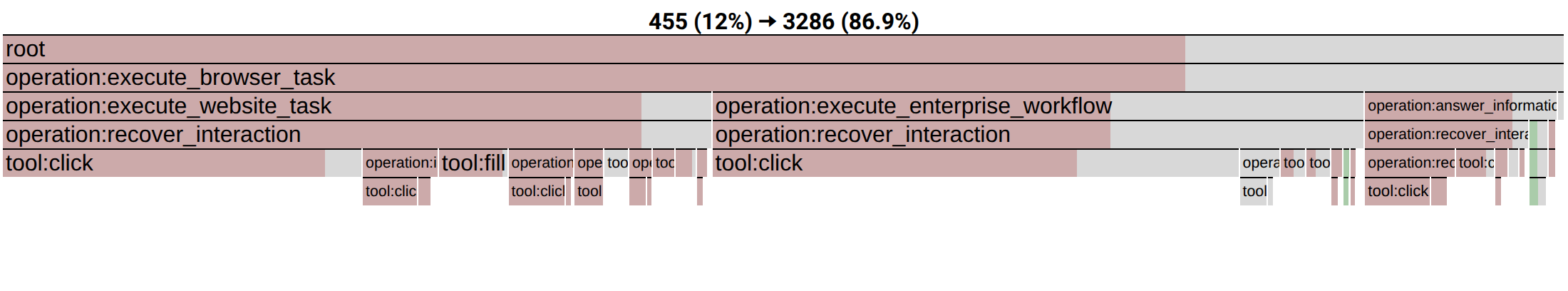}\\[2pt]
\textbf{(b) Completion focus}\\[-2pt]
\includegraphics[width=\columnwidth]{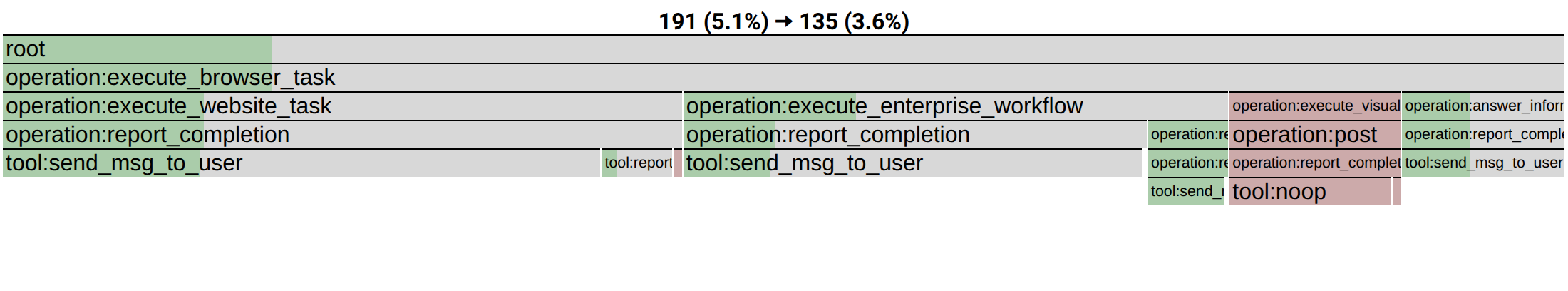}
\caption{Stock-pprof differential flame graphs (bad minus good). Box width sums both contributions; inset shows net difference (rose: bad excess; green: good excess). Recovery dominates bad runs; completion favors good runs.}
\Description{Two stock-pprof differential flame graphs. The recovery panel is
dominated by rose bad-side growth beneath browser-task responsibilities. The
completion panel has a green good-side excess through report-completion and
send-message paths.}
\label{fig:diff-flamegraph}
\end{figure}

\paragraph{Use case 3: what do failing runs do differently?}
A team has 440 web-agent runs~\cite{agentrewardbench} over 125 tasks, each with successful and failed attempts (202 successful, 238 failed); reading all transcripts (7{,}229 operations) is infeasible.
We pair failed and successful runs per task (338 occurrences), profile each group, and subtract them (failed minus successful) into a signed difference profile (Figure~\ref{fig:diff-flamegraph}). The result is immediate: failed runs spend 44.6\% of steps in \texttt{recover interaction} (retrying, re-searching, re-navigating) versus 12.0\% for successful runs.
The hierarchy decomposes recovery into verification challenges, repeated searches, mistaken navigation, and record retries; source labels identify sessions contributing each width.
Failing agents thus get stuck in retry loops instead of backing out to try another approach.
This structure matches independent expert looping labels on 435 trajectories at AP .634 versus a .398 random baseline (difference interval [.181, .293]), confirming a real behavioral pattern. A fixed-chain repeated/error control reaches the same detection quality (AP .656), but only the recursive profile provides this decomposition and source navigation.

\begin{figure}[tb]
\centering
\includegraphics[width=0.94\columnwidth]{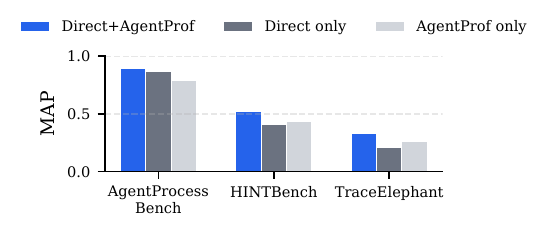}
\caption{MAP over the complete RQ2 datasets (614, 400, and 220 queries included in MAP). Higher is better.}
\Description{A grouped bar chart showing MAP scores for three methods across three workloads. Direct+\sys performs best on all workloads.}
\label{fig:rq2-map}
\end{figure}

\paragraph{Supplementing benchmark's own diagnostics.}
\emph{Setup.} Each MAP query has one or more annotated faulty operations.
A method ranks its operations, scored by average precision (AP equals
reciprocal rank for one faulty operation) and averaged as MAP~\cite{robertson2008ap}.
We run complete AgentProcessBench, TraceElephant, and HINTBench workloads (the
complete released test snapshot, 536 of 629 paper-reported trajectories); all
profiler outputs precede fault-annotation disclosure.
We average per-query AP over the 614, 400, and 220 MAP queries; 522 trajectories without annotated faulty operations provide dataset coverage but are excluded from MAP.
The \emph{direct diagnostic} is each benchmark's per-operation judge output, including LLM-as-judge methods, which reads the full trace and reference answer before naming the responsible agent and decisive step.
\emph{Direct-only} uses only that diagnostic; \emph{Direct+\sys} breaks ties by group score (vote mean for AgentProcessBench; otherwise 95\% Wilson bound), with HINTBench field order fixed on 80 validation trajectories;
\emph{\sys-only} uses only the profiler (Figure~\ref{fig:rq2-map}).

\emph{Results.} Direct+\sys beats Direct-only by 0.031, 0.107, and 0.117 MAP across three workloads (all statistically significant\appnote{; 95\% intervals in Appendix~\ref{app:rq2}}; Figure~\ref{fig:rq2-map}).
Evidence-preserving grouping drives the localization gain. Semantic naming enables cross-run attribution (RQ1) and concentrates attention (measured below).

\paragraph{Profile-guided reading on TraceElephant.}
On 220 queries, a Grok 4.5~\cite{grok45} agent-as-judge ranks operations by fault likelihood.
Full-trace reading reaches MAP .502 at 12.6K tokens/query. A profile-guided reader selects at most five groups, reaching .455 while opening 53\% of source, versus .465 and 65\% with raw-action names\appnote{intervals in Appendix~\ref{app:rq2}}. Per-query full reading is also context-window bounded.
\emph{Takeaway.} Profilers complement debuggers: profiles answer cross-run questions and guide inspection to relevant content. Profiled failure structure matches independent human labels across 435 trajectories. Profiles supplement existing judges' localization signal, while semantic names concentrate agent attention, reducing opened source characters from 65\% to 53\%.

\paragraph{Use case 4: guiding a real repair.}
This case tests whether a profile finding can guide effective repair.
A standard \sys profile of eight ToolSandbox scenarios isolates recurring
call-ID syntax failure (5/21 tool operations).
A profile-only analyst---given pprof output but no raw traces---identifies the
compatibility-layer boundary as the repair target.
The resulting one-line fix removes all syntax failures and, on 23 held-out
confirmation scenarios (69 BEFORE/repair pairs), reduces agent-model tokens
by 19.0\% while passing a fixed $-.05$ official-similarity quality threshold
\appnote{details in Appendix~\ref{app:rq2-repair}}.

\subsection{RQ3: How Accurate Are the Identifiers?}

\emph{Setup.} We test whether recursive segmentation recovers human-recognizable task structure against 2{,}948 human-annotated stages from all 405 CodeTraceBench trajectories.
F1~\cite{bagga-baldwin-1998-entity-based} asks whether related operations
are grouped; exact adjacent-boundary F1~\cite{ruokolainen2016segmentation}
asks whether cuts match human cuts. Baselines receive identical inputs
(Table~\ref{tab:rq3-codetrace}).
Each output is evaluated at its predicted level: literal names use
accuracy and macro-F1~\cite{lewis2004rcv1}, permutation-invariant partitions
V-measure~\cite{rosenberg-hirschberg-2007-v-measure} or ordinary B$^3$,
and adjacent interval boundaries exact precision, recall, and F1. For
legacy field-valued datasets, conversion maps each predicted group into
the same interval-annotation format before \sys folds the
original additive weights.
Codex~\cite{codex} receives each trajectory (prompts, commands, and output summaries; no visible labels or scores) and emits sparse interval marks over 17{,}148 steps without stage annotations (loaded after fixing all 405 outputs): 4{,}496 marks at semantic depths one (3), two (2{,}873), three (1{,}588), or four (32), with no requested depth.
Name normalization maps 3{,}895 free-form interval names to 783 stable short names (e.g., \texttt{diagnose authentication}) with zero adjacent display-path collisions, rejecting unresolved names. Identical cross-session names merge, folding identical paths in the final profile.

\begin{table}[tb]
\centering\small
\setlength{\tabcolsep}{2.0pt}
\begin{tabular}{@{}lrrrr@{}}
\toprule
Method & B$^3$ P & B$^3$ R & B$^3$ F1 & Bound. F1 \\
\midrule
Codex segmentation & 0.793 & 0.736 & \textbf{0.764} & \textbf{0.480} \\
Statistical recurrence & 0.782 & 0.575 & 0.663 & 0.266 \\
Causal Qwen2.5-3B task stack & 0.557 & 0.792 & 0.650 & 0.257 \\
Raw-action grouping & 0.891 & 0.388 & 0.541 & --- \\
Native source tree & 0.975 & 0.249 & 0.397 & 0.259 \\
Source-native step & 0.983 & 0.221 & 0.361 & 0.246 \\
\bottomrule
\end{tabular}
\caption{Agreement with human stages on all 405 CodeTraceBench trajectories.}
\label{tab:rq3-codetrace}
\end{table}

\emph{Results.} Codex segmentation reaches 0.764 B$^3$ F1 and 0.480
boundary F1, beating the strongest automatic baseline (multi-resolution recurrence) by +0.101 and +0.214\appnote{Appendix~\ref{app:rq3} intervals}.
A stateful per-turn Qwen2.5-3B baseline reaches 0.650 B$^3$ F1, showing meaningful segmentation by smaller models.
The marks exactly conserve 20{,}866 operations and
494{,}862{,}929 tokens.
Predicted groups remain largely pure gold-stage subsets
(B$^3$ precision 0.793), with residual error from over-segmentation
rather than missed transitions: disagreements subdivide work instead of merging unrelated responsibilities,
preserving per-group coherence\appnote{breakdown in Appendix~\ref{app:rq3}}.
The interface generalizes across method families and segmentation scales: on OSWorld-Human~\cite{osworldhuman}, supervised Naive Bayes reaches 0.816 B$^3$ F1, a no-LLM recurrence segmenter 0.786, and the unchanged Codex instruction uses coarser task-level boundaries (0.448 B$^3$ F1). 
For closed-label literal tags, Qwen3.6-27B~\cite{qwen36} labels 1{,}012 AgentBoard goals~\cite{agentboard} at 0.695 macro-F1 and 2{,}737 AutoCodeRover, OpenHands, and RepairAgent trajectory actions~\cite{traceview-2026,bouzenia-pradel-2025-trajectories} at 0.498 macro-F1. 
\emph{Takeaway.} Recursive segmentation recovers human-recognizable task structure: 0.764 B$^3$ F1 against human stages, +0.101 over the strongest automatic baseline. Non-LLM implementations of the same interface remain viable without a model.

\subsection{RQ4: What Is the Profiling Cost?}

\emph{Setup.} We test profiling practicality through segmentation time and tokens plus profile replay speed. Cost comprises one automatic segmentation pass per corpus, then deterministic construction/replay.

\emph{Results.} Codex (GPT-5.6-sol-high) segments all 405 CodeTraceBench trajectories in 37 minutes with up to four workers, averaging 29{,}754 input and 573 output tokens per trajectory.
With fixed marks, construction scales linearly: the 27{,}765-operation union completes in 1.16\,s at 465\,MiB peak RSS, only 5.25\,MiB (1.14\%) above raw-action grouping\appnote{details in Appendix~\ref{app:rq4-cost}}.
Deterministic construction of all 440 trajectories takes 0.26\,s (operations) and 0.25\,s (tokens): segmentation dominates cost, while construction remains sub-second.

\paragraph{Reducing annotation cost.}
Automatic annotation dominates construction. A compact skeleton plus
selected full results (SELECTIVE), instead of every turn's
complete content (FULL) reduces provider tokens 20.4\% on 32 held-out
CodeTraceBench task clusters while retaining full operation coverage and
meeting a fixed $-.03$ quality threshold on B$^3$ and boundary F1
\appnote{details in Appendix~\ref{app:rq4-selective}}.

\emph{Takeaway.} Segmentation is one-time (37 min for 405 trajectories);
each later view, measure switch, and inspection costs about one second.
Selective evidence cuts annotation tokens 20.4\% with complete coverage
and preserved structural quality.

\section{Related Work}

Classic profilers fold runtime call stacks over stable code identity, and Pivot Tracing groups causally related measurements~\cite{pprof,flamegraphs,pivottracing}.
Agent observability platforms such as LangSmith, Langfuse, and Phoenix~\cite{langsmith,langfuse,phoenix} record spans with metadata for per-execution debugging, and analytics layers such as Datadog Patterns and LangSmith Insights~\cite{datadog-patterns,langsmith-insights} roll up cross-trace hierarchies or input distributions, but none constructs recursive semantic responsibility with conserved additive measures in a standard profile.
Cross-run analyses (TraceProbe, Graphectory, Act$\cdot$onomy, CHIEF, Hodoscope, TraceGraph)~\cite{traceprobe2026,graphectory2026,actonomy2026,chief2026,hodoscope2026,tracegraph2026} and per-run diagnosis and localization~\cite{agentrx,telbench,agentatlas,trajad,agentfixer,li2026codetracer,agentprocessbench,hintbench,traceelephant,mpbench2026} answer what an agent did and which step went wrong.
\sys instead asks how much of an additive resource a task or workflow phase is responsible for. It recovers variable-depth responsibility from source content (where Act$\cdot$onomy fixes three levels and TraceProbe one), conserves measures exactly across count, token, and time widths, keeps LLM/tool calls as leaf nodes, and emits standard pprof. No compared system provides all four, and \sys stays complementary to per-run diagnosis, as RQ2 measures directly.

\section{Conclusion}

Agent observability needs profiling, not only debugging.
\sys shows that the semantic operation stack model, driven by
recursive operation segmentation, brings hierarchical attribution to
agent trajectories, effectively attributes resources, locates problems, and helps optimize token cost at practical profiling cost.
Adaptive triggering, intent-level state-transition models,
profile-driven runtime control, and multi-project evaluation with
tracing-ecosystem integration remain future work.

\bibliographystyle{ACM-Reference-Format}
\bibliography{references}

\end{document}

%% file: figures/fig-architecture.tex
\begin{tikzpicture}[
  node distance=0.5cm and 0.8cm,
  box/.style={rectangle, draw, fill=blue!15, minimum width=2.2cm,
              minimum height=0.8cm, font=\small, align=center,
              rounded corners=2pt},
  input/.style={rectangle, draw, fill=gray!15, minimum width=2.0cm,
               minimum height=0.7cm, font=\small, align=center,
               rounded corners=2pt},
  output/.style={rectangle, draw, fill=green!12, minimum width=1.6cm,
                minimum height=0.7cm, font=\small, align=center,
                rounded corners=2pt},
  arr/.style={-{Stealth[length=1.8mm]}, thick, gray!70},
]

\node[input] (traj) {Agent\\Trajectories};

\node[box, right=0.8cm of traj, fill=orange!15] (intent) {Operation\\Segmentation};
\node[font=\footnotesize, below=0.05cm of intent, text=gray] (backends) {once per trajectory};

\node[box, right=0.8cm of intent] (stack) {Projection};
\node[font=\footnotesize, below=0.05cm of stack, text=gray] {query time};

\node[box, right=0.8cm of stack, fill=green!12] (out) {Output};
\node[font=\footnotesize, below=0.05cm of out, text=gray] {pprof};

\draw[arr] (traj) -- (intent);
\draw[arr] (intent) -- (stack);
\draw[arr] (stack) -- (out);

\end{tikzpicture}